\documentclass[sigconf,nonacm]{acmart}
\usepackage{pifont}
\usepackage{amssymb}
\usepackage{enumitem}
\copyrightyear{2020}
\acmYear{2020}
\setcopyright{rightsretained}
\acmConference[HAI '20]{Proceedings of the 8th International Conference on Human-Agent Interaction}{November 10--13, 2020}{Virtual Event, NSW, Australia}
\acmBooktitle{Proceedings of the 8th International Conference on Human-Agent Interaction (HAI '20), November 10--13, 2020, Virtual Event, NSW, Australia}
\acmDOI{10.1145/3406499.3418760}
\acmISBN{978-1-4503-8054-6/20/11}

\begin{document}

\fancyhead{}

\title{SEAN: Social Environment for Autonomous Navigation}

\author{N. Tsoi}
\affiliation{%
  \institution{Yale University}
  \streetaddress{51 Prospect St}
  \postcode{06511}
}
\email{nathan.tsoi@yale.edu}

\author{M. Hussein}
\affiliation{%
  \institution{Rutgers University}
}

\author{J. Espinoza}
\affiliation{%
  \institution{Yale University}
  \streetaddress{51 Prospect St}
}

\author{X. Ruiz}
\affiliation{%
  \institution{Yale University}
}

\author{M. V\'azquez}
\affiliation{%
  \institution{Yale University}
}

\renewcommand{\shortauthors}{Tsoi et al.}


\begin{abstract}

Social navigation research is performed on a variety of robotic platforms, scenarios, and environments. Making comparisons between navigation algorithms is challenging because of the effort involved in building these systems and the diversity of platforms used by the community; nonetheless, evaluation is critical to understanding progress in the field. In a step towards reproducible evaluation of social navigation algorithms, we propose the Social Environment for Autonomous Navigation (SEAN). SEAN is a high visual fidelity, open source, and extensible social navigation simulation platform which includes a toolkit for evaluation of navigation algorithms. We demonstrate SEAN and its evaluation toolkit in two environments with dynamic pedestrians and using two different robots.

\end{abstract}

\maketitle

\section{Introduction}

\begin{figure}[t]
    \centering
    \begin{minipage}{.24\textwidth}
        \centering
        \includegraphics[width=0.99\linewidth]{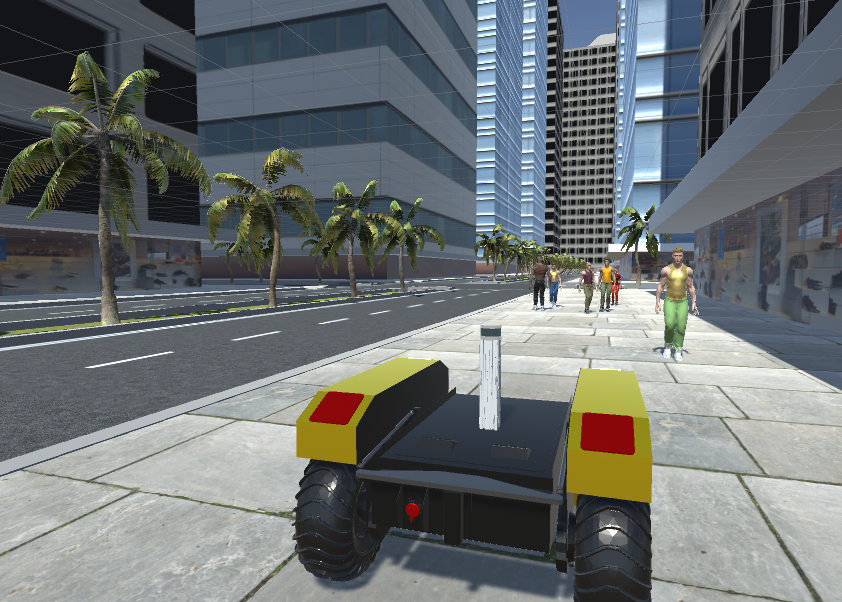}
    \end{minipage}%
    \begin{minipage}{0.24\textwidth}
        \centering
        \includegraphics[width=0.99\linewidth]{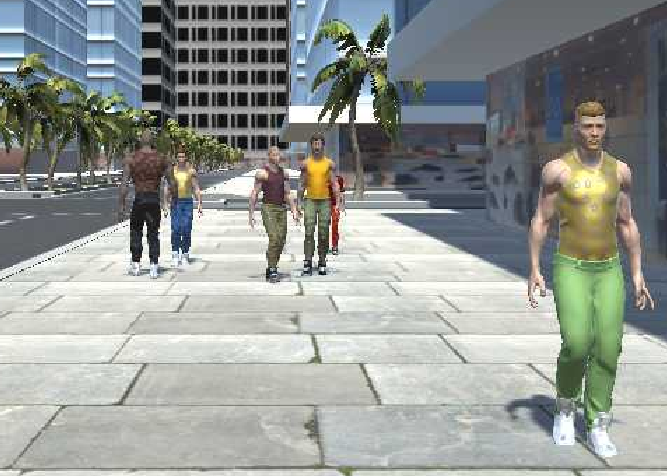}
    \end{minipage}
    \centering
    \begin{minipage}{.24\textwidth}
        \centering
        \includegraphics[width=0.99\linewidth]{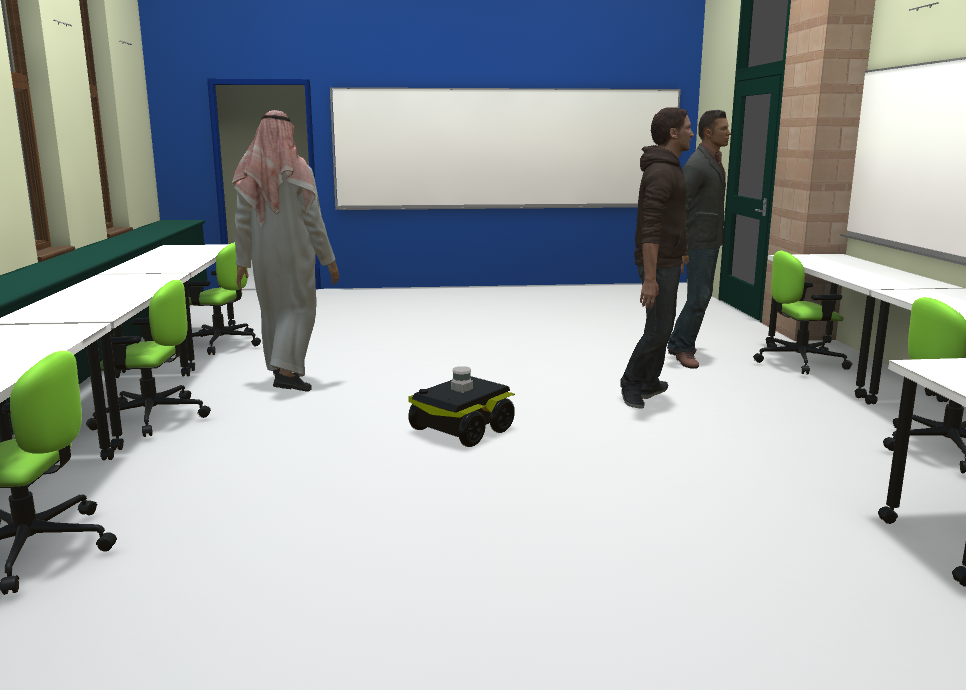}
    \end{minipage}%
    \begin{minipage}{0.24\textwidth}
        \centering
        \includegraphics[width=0.99\linewidth]{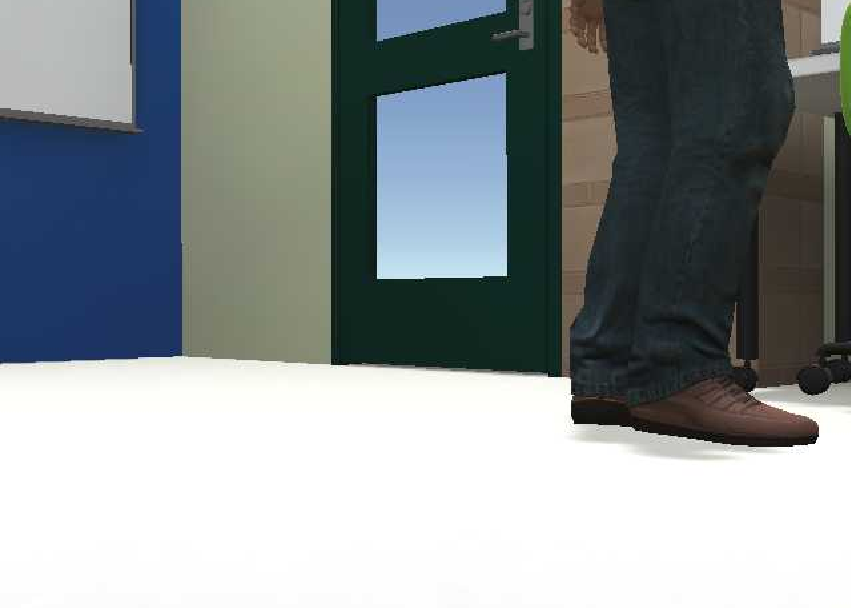}
    \end{minipage}
    \caption{SEAN's rendering a world (left) and view of the scene from the robot's perspective (right). The outdoor city scene (top) and the lab scene (bottom) include dynamic pedestrians for studying social robot navigation.}
    \label{fig:scenes}
\end{figure}

Simulation is useful along the whole development cycle of robotic systems including data collection, features development, testing, and deployment \cite{ros2016synthia, nourbakhsh2005human}. Simulation is key for the verification of safety-critical systems and is particularly relevant for companies that make robots for mainstream audiences \cite{park2013simulation}. 

Driven by the gaming industry and demand for autonomous vehicles, the robotics community has recently experienced a rapid increase in the quality and features available in simulation tools. These advancements led to simulation environments for self-driving vehicles \cite{Dosovitskiy17,amini2020learning} and aerial vehicles \cite{meng2015ros+, airsim2017fsr, guerra2019flightgoggles}. Crowd simulations have improved as well \cite{tai2018socially,curtis2016menge, aroor2017}, although often independently of simulation environments for mobile robots, including environments that build on game engines \cite{hussein2018ros,konrad2019simulation}, or state-of-the-art rendering like Gibson\cite{xiazamirhe2018gibsonenv} or ISAAC.\footnote{https://developer.nvidia.com/isaac-sim} This disconnect has led to a gap in high fidelity simulation environments for evaluating social robot navigation in pedestrian settings, e.g., service robots that need to operate nearby people and subject to social conventions. Our work, depicted in Figure \ref{fig:scenes}, is a step towards filling this gap. 

We propose SEAN, a Social Environment for Autonomous Navigation, as an extensible and open source simulation platform. SEAN includes animated human characters useful for studying human-robot social interactions in the context of  navigation. As other recent simulators \cite{meng2015ros+,airsim2017fsr,hussein2018ros,konrad2019simulation}, SEAN leverages modern graphics and physics modeling tools from the gaming industry, providing a flexible development environment in comparison to more traditional robotics simulators like Gazebo \cite{koenig2013many}. We provide two ready-to-use scenes with components that allow social agents to navigate according to standard pedestrian models. We provide integration with the Robot Operating System (ROS), which allows for compatibility with existing navigation software stacks. An important contribution of this work is a toolkit for repeated execution of navigation tasks and logging of navigation metrics. We hope SEAN facilitates developing, testing and evaluating social navigation algorithms in the future.


\section{Simulation Platform}

SEAN is a collection of tools in the Unity 3D game engine\footnote{https://unity.com/} and the Robot Operating System (ROS) \cite{quigley2009ros} that allows for control of a mobile robot in a dynamic, simulated human environment. Unity implements the NVIDIA PhysX physics engine, which has been found to provide promising results for robot simulation \cite{konrad2019simulation}. Communication between ROS and Unity is implemented as a set of scripts executed as part of the Unity scripting run-time model and implemented via the ROS\# library.\footnote{https://github.com/siemens/ros-sharp} SEAN's architecture balances between a) ease of integration with navigation systems (or robot teleoperation) via ROS, b) high visual fidelity for creating immersive environments and enable vision-based navigation methods, and c) a cross-platform ecosystem that supports iterative development. SEAN works in Windows 10 and Ubuntu 18.04 with ROS Melodic.

The key tenets of our approach are usability and flexibility. While these often seem at odds, we seek these goals by providing a set of scenes, robots, and evaluation metrics within the platform to enable users to use the system with minimal preliminary work. Additionally, we maintain an open source repository and supporting documentation to allow the community to improve our social navigation environment.\footnote{https://sean.interactive-machines.com/}. Our contributions are an effort to begin to explore the challenging problem of fairly and reproducibly benchmarking algorithms for human-robot social interactions. 

\textbf{Scenes}: A scene is a 3D environment in which a robot operates. With our initial release, we provide a high-fidelity model of a lab environment and a larger outdoor city scene (Fig. \ref{fig:scenes}). Because humans play a key part in the study of robot navigation in these environments, for each scene we have created reasonable start and goal positions for human agents to navigate. To this end, SEAN uses a combination of crowd flow prediction  \cite{sohn2020} and Unity's built-in path planning algorithm. The system is parameterized such that we can easily deploy an appropriate number of agents given the size and context of the scene. We can also vary the density of pedestrians across experiments in a repeatable manner. SEAN's online documentation explains how to create and modify scenes.

\textbf{Robots and Sensors}: SEAN provides 2 robot models ready to run: a medium size Clearpath  Jackal, which is suitable for indoor and flat outdoor environments; and a Warthog with 254mm of ground clearance. The Warthog is more suitable for outdoor environments due to its bigger size (Fig. \ref{fig:scenes}). Because neither robot comes equipped with standard sensors, we outfitted them with a simulated Velodyne VLP-16,\footnote{https://github.com/Field-Robotics-Japan/unit04\_unity/} a LIDAR scanner, and a simulated RGB camera.

\textbf{Evaluation Toolkit:} SEAN's toolkit for evaluating social navigation algorithms centers on the Trial Runner, which enables repeatable and automatic execution of navigation tasks. The Trial Runner performs a \textit{trial} by executing a collection of  point-to-point navigation \textit{episodes}. Each episode begins with the Trial runner configuring the scene, actors, and robot positions. Pedestrians are assigned goal positions and a ROS navigation goal is used to indicate the desired final pose for the robot. As the robot navigates, the Trial Runner records relevant metrics. It starts a new episode once the robot has moved to a sufficiently close location to the destination or the episode times out. While the initial conditions for each episode are random by default, they are recorded at the beginning of an episode. This allows to replay the episode for fair comparisons of   navigation methods.

SEAN currently tracks the following navigation metrics: whether or not the robot reached the goal position, how long it took to reach the final position, collisions with static objects, and the robot's final distance to the goal position. The latter metric is particularly useful for comparison in challenging tasks. In addition, SEAN can continuously track metrics related to social interactions. Currently, we track the closest distance between the robot and pedestrians, as well as the number of collisions with pedestrians, which are recorded separately from collisions with all other objects. 
These metrics are common in the social navigation literature \cite{trautman2015robot,okal2016learning,tai2018socially,mavrogiannis2019effects,pokle2019deep} and serve as a starting point for comparisons among navigation approaches. We plan to expand this set of metrics in the future.

\begin{table}[t!p]
    \centering
    \caption{Sample Jackal (J) robot and Warthog (W) robot results. Controlled via the ROS Navigation Stack or teleoperated as indicated by \textbf{*}. $\mu \pm \sigma$ over 10 episodes. }
    \resizebox{\columnwidth}{!}{%
        \begin{tabular}{cc|ccccc}
            Scene & Robot & Elapsed (sec.) & Complete & Final Dist (m) & Ped. Dist (m) & Collisions \\ \hline
            Lab & J & $24.51 \pm 19.36$ & $60\%$ & $2.26 \pm 2.92$m & $1.54 \pm 1.76$m & $7.1 \pm 9.4$\\ \hline
            Lab & J * & $21.6 \pm 28.08$ & $88\%$ & $1.14 \pm 1.99$ & $0.92 \pm 1.16$ & $4.63 \pm 5.83$\\ \hline
            City & J & $37.09 \pm 13.74$ & $29\%$ & $9.54 \pm 8.94$ & $0.64 \pm 0.42$ & $20 \pm 30.83$ \\ \hline
            City & J * & $38.54 \pm 29.5$ & $80\%$ & $4.59 \pm 11.87$ & $1.06 \pm 0.67$ & $3.1 \pm 7.58$ \\ \hline
            City & W * & $31.7 \pm 20.94$ & $100\%$ & $0.48 \pm 0.01$ & $2.27 \pm 1.08$ & $0 \pm 0$
        \end{tabular}
    }
    \label{tbl:example-results}
\end{table}

Table \ref{tbl:example-results} provides example results by the Trial Runner for the ROS Navigation Stack \cite{guimaraes2016ros}, which was minimally tuned, and a teleoperated robot. Localization for the ROS Nav. Stack was performed via SLAM \cite{grisetti2007improved}. Low performance is attributed to not taking into account human actors during mapping and overly conservative navigation behavior in the dynamic environments \cite{trautman2015robot}. 

Teleoperation was implemented through a ROS node that connected to a gamepad controller. The teleoperated Jackal did not reach 100\% of the target goals because people blocked its way and the episodes timed out. Nonetheless, teleoperation was an interesting baseline for automated methods. It can also serve  to gather demonstrations or human preferences for navigation trajectories in the future \cite{kuderer2013teaching,vasquez2014inverse}.

As shown by the examples, SEAN is a valuable tool to systematically evaluate performance. Additionally, SEAN can accelerate the development of navigation systems by helping identify problems early on. We do not claim, though, that simulation is a replacement for real-world testing; instead, SEAN compliments  real-world evaluation of human-robot interactions in the context of navigation.

\section{Conclusion \& Future Work}

We presented SEAN, a collection of open-source tools and social environments for autonomous robot navigation. Our hope is that the flexibility of the platform encourages researchers to  make more comparisons between social robot navigation methods, and more easily study human-robot interactions in the future. For instance, videos from SEAN could be used for qualitative human evaluation of navigation approaches \cite{tjomsland2020mind}. While we currently provide a limited number of scenes, robots, and evaluation metrics, we plan to expand these features in the future. We also plan to test SEAN in terms of sim-to-real transfer of navigation policies and human evaluation of robot behaviors. 

\section{Acknowledgements}

This work was funded by a 2019 Amazon Research Award and partially supported by the National Science Foundation (NSF) under Grant No. 1910565. Any opinions, findings, and conclusions or recommendations expressed in this material are those of the author(s) and do not necessarily reflect the views of Amazon nor the NSF.

\bibliographystyle{ACM-Reference-Format}
\balance
\bibliography{references}

\end{document}